\documentclass[letterpaper,twocolumn,10pt]{article}
\usepackage{usenix,epsfig,endnotes}
\usepackage{algorithm}
\usepackage{algpseudocode}
\usepackage{amsmath}
\usepackage{amssymb}
\usepackage{booktabs}
\usepackage{graphicx}
\usepackage{float}
\usepackage{pgfplots}
\pgfplotsset{compat=1.18}
\usepackage{caption}
\usepackage{subcaption}
\graphicspath{{./AIC_assets/}}
\usepackage{enumitem}
\usepackage{listings}
\usepackage{xcolor}
\usepackage{tikz}
\usepackage{adjustbox}
\usepackage{hyphenat}

\begin{document}

\date{}

\title{\Large \bf AIConfigurator: Lightning-Fast Configuration Optimization for Multi-Framework LLM Serving}

\author{
{\rm Tianhao Xu} \quad {\rm Yiming Liu} \quad {\rm Xianglong Lu} \quad {\rm Yijia Zhao} \quad {\rm Xuting Zhou} \quad {\rm Aichen Feng} \quad {\rm Yiyi Chen} \\[0.5em]
{\rm Yi Shen} \quad {\rm Qin Zhou} \quad {\rm Xumeng Chen} \quad {\rm Ilya Sherstyuk} \quad {\rm Haorui Li} \quad {\rm Rishi Thakkar} \quad {\rm Ben Hamm} \\[0.5em]
{\rm Yuanzhe Li} \quad {\rm Xue Huang} \quad {\rm Wenpeng Wu} \quad {\rm Anish Shanbhag} \quad {\rm Harry Kim} \quad {\rm Chuan Chen} \quad {\rm Junjie Lai} \\[0.5em]
NVIDIA \\
}

\maketitle

\thispagestyle{empty}

\section*{Abstract}
\vspace{-1em}
Optimizing Large Language Model (LLM) inference in production systems is increasingly difficult due to dynamic workloads, stringent latency/throughput targets, and a rapidly expanding configuration space. This complexity spans not only distributed parallelism strategies (tensor/pipeline/expert) but also intricate framework-specific runtime parameters such as those concerning the enablement of CUDA graphs, available KV-cache memory fractions, and maximum token capacity, which drastically impact performance. The diversity of modern inference frameworks (e.g., TRT-LLM, vLLM, SGLang), each employing distinct kernels and execution policies, makes manual tuning both framework-specific and computationally prohibitive.
We present AIConfigurator, a unified performance-modeling system that enables rapid, framework-agnostic inference configuration search without requiring GPU-based profiling. AIConfigurator combines (1) a methodology that decomposes inference into analytically modelable primitives—GEMM, attention, communication, and memory operations while capturing framework-specific scheduling dynamics; (2) a calibrated kernel-level performance database for these primitives across a wide range of hardware platforms and popular open-weights models (GPT-OSS, Qwen, DeepSeek, LLama, Mistral); and (3) an abstraction layer that automatically resolves optimal launch parameters for the target backend, seamlessly integrating into production-grade orchestration systems.
Evaluation on production LLM serving workloads demonstrates that AIConfigurator identifies superior serving configurations that improve performance by up to 40\% for dense models (e.g., Qwen3-32B) and 50\% for MoE architectures (e.g., DeepSeek-V3), while completing searches within 30 seconds on average, enabling the rapid exploration of vast design spaces—from cluster topology down to engine specific flags.

\section{Introduction}
The rapid evolution of Large Language Models (LLMs) has placed unprecedented demands on computational infrastructure. As state-of-the-art parameter counts scale from hundreds of millions to hundreds of billions, the efficiency of inference deployment has become a critical determinant of economic viability. However, optimizing these deployments is fraught with complexity. Service providers must navigate a combinatorial explosion of configuration parameters—ranging from tensor, pipeline, and expert parallelism strategies to granular settings for batch sizes and quantization. Traditional performance tuning, often reliant on manual benchmarking and exhaustive testing, is increasingly untenable. With the rising cost of modern GPUs, manual exhaustive testing becomes prohibitively expensive. These methods require significant engineering effort to converge on solutions that frequently remain sub-optimal, leaving substantial performance potential untapped.

The emergence of disaggregated serving~\cite{zhong2024distservedisaggregatingprefilldecoding, qin2024mooncake}—separating prefill and decode phases onto distinct compute resources—has further complicated this landscape. While disaggregation promises to optimize "Goodput" (throughput under strict latency constraints), it is not a universally superior solution; gains depend heavily on the interplay between model architecture, hardware topology, and network bandwidth. Practitioners face a difficult trade-off: does the scheduling flexibility of disaggregation outweigh the communication overhead for a specific workload? Furthermore, configuring such systems to satisfy rigorous Service Level Agreements (SLAs)—specifically Time-To-First-Token (TTFT) and Time-Per-Output-Token (TPOT)—creates a design space that can easily exceed 10,000 permutations.

Beyond architectural decisions, the complexity is further compounded by the intricacies of the inference engines themselves. Modern frameworks such as TensorRT-LLM~\cite{githubGitHubNVIDIATensorRTLLM}, vLLM~\cite{kwon2023efficient}, and SGLang~\cite{zheng2023efficiently} expose a myriad of tunable runtime flags, such as those pertaining to the enablement of CUDA graphs, available KV-cache memory fractions, and maximum token capacity, that drastically impact performance. A configuration that maximizes throughput for a specific workload often proves brittle or inefficient as the serving environment evolves. Consequently, developers frequently abandon the tuning process, defaulting to conservative settings that leave significant performance potential untapped.

While black-box optimization methods such as Vizier~\cite{golovin2017google} or automated serving frameworks like Morphling~\cite{wang2021morphling} can help speed up finding the optimal configs, they still require a substantial amount of GPU hours to converge on a solution for each specific scenario.

To address these challenges, we present AIConfigurator, a specialized toolkit designed to optimize LLM inference across diverse inference frameworks. Unlike generic simulators reliant on theoretical abstractions, AIConfigurator employs a data-driven approach rooted in operation-level performance modeling. By decomposing inference into fundamental kernels—such as GEMM computations, attention mechanisms, and communication primitives (e.g., all-reduce, P2P)—and utilizing interpolation of real system data, the toolkit achieves high-fidelity estimates tailored to NVIDIA platforms (Ampere, Ada, Hopper, and Blackwell). Our contributions include:
\begin{itemize}
\item Designing a system capable of navigating the complex inference configuration space to identify optimal settings in seconds with high precision.
\item Demonstrating the effectiveness of the system through seamless integration with mainstream inference frameworks—including vLLM, SGLang, TRTLLM, and NVIDIA’s Dynamo—delivering actionable, production-ready recommendations.
\item Conducting a comprehensive evaluation by benchmarking against ground-truth silicon data.
\end{itemize}

\begin{figure*}[t]
    \centering
    \includegraphics[width=0.8\textwidth]{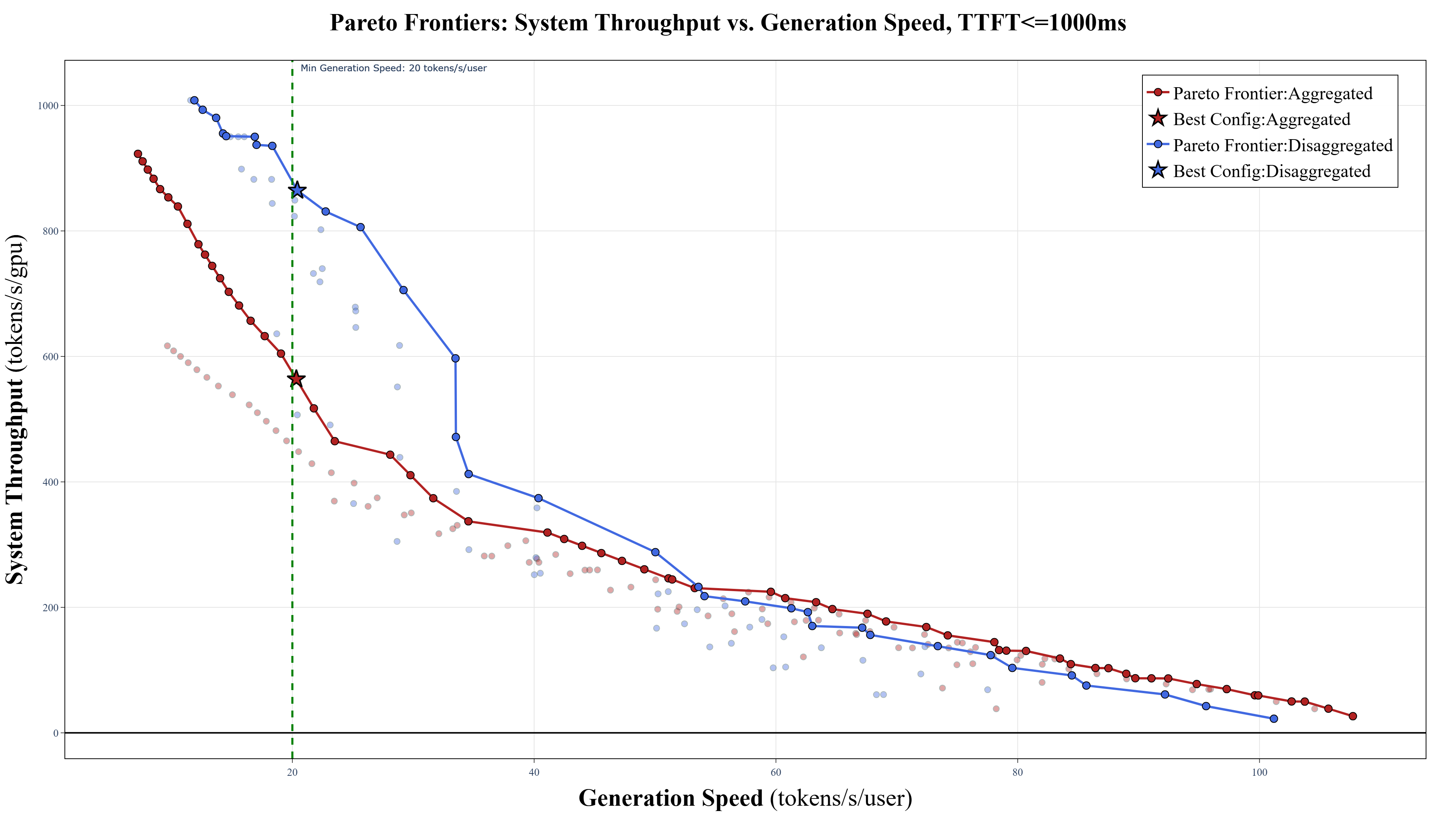}
    \caption{\small AIConfigurator projected Throughput vs Speed Pareto frontiers for Qwen3-235B running on 64 H200 GPUs. All serving configurations that can achieve a TTFT (Time to First Token) $\leq$ 1000ms are plotted on the chart.}
    \label{fig:pareto_chart}
    \vspace{-2.5mm}
\end{figure*}

\section{Background}
\subsection{LLM Inference Optimization}
LLM inference optimization involves three interdependent pillars. \emph{Advanced scheduling}---including continuous batching~\cite{yu2022orca}, PagedAttention~\cite{kwon2023efficient}, chunked prefills~\cite{agrawal2024taming}, and disaggregated serving~\cite{zhong2024distservedisaggregatingprefilldecoding}---maximizes hardware utilization by addressing the distinct computational profiles of prefill (compute-bound) and decode (memory-bound) phases. \emph{Model parallelism} distributes large models across GPUs via Tensor Parallelism (TP)~\cite{shoeybi2019megatron}, Pipeline Parallelism (PP)~\cite{huang2019gpipe}, and Expert Parallelism (EP)~\cite{liu2024deepseek} for MoE architectures. \emph{Configuration tuning} navigates the resulting combinatorial space; while simulators like Vidur~\cite{agrawal2024vidur} and APEX~\cite{lin2024apex} enable rapid exploration, their reliance on theoretical roofline models often diverges from production performance.

\subsection{Aggregated vs.\ Disaggregated Serving}
Disaggregated serving~\cite{zhong2024distservedisaggregatingprefilldecoding, qin2024mooncake} separates prefill and decode onto distinct GPU pools, enabling independent scaling but introducing KV-cache transfer overhead. Splitwise~\cite{patel2024splitwise} shows this overhead can negate benefits for short contexts. The optimal architecture depends on workload mix (prefill-heavy vs.\ decode-heavy), interconnect bandwidth, and cluster scale---aggregated serving with chunked prefills often outperforms disaggregation for smaller deployments. This complexity motivates AIConfigurator, which models both architectures to identify optimal configurations.

\section{Motivation}
Traditional heuristics like "TP within node, PP across nodes" fail to capture non-linear interactions between compute and network bandwidth. Studies~\cite{zheng2022alpa, miao2023galvatron} show automated search can outperform manual tuning by $>$2$\times$ in cost-efficiency.

\textbf{Framework Heterogeneity.} Production inference spans diverse frameworks with distinct performance characteristics: \textbf{vLLM}~\cite{kwon2023efficient} (PagedAttention, Python-based scheduling), \textbf{SGLang}~\cite{zheng2023efficiently} (RadixAttention, Triton kernels), \textbf{TensorRT-LLM}~\cite{githubGitHubNVIDIATensorRTLLM} (static graph optimization, custom kernels), and \textbf{NVIDIA Dynamo}~\cite{githubGitHubAidynamodynamo} (backend-agnostic orchestration). Each exhibits unique performance cliffs governed by a myriad of framework-specific runtime flags that generic models cannot effectively capture.

\textbf{Configuration Tuning Gap.} Current approaches---nightly benchmarks~\cite{semianalysisInferenceMAXSemiAnalysis} and curated recipes~\cite{githubGitHubVllmprojectrecipes}---are static lookup tables insufficient for dynamic production environments. AIConfigurator provides algorithmic search that identifies SLA-compliant configurations across the multi-dimensional space of parallelism, batch sizes, and serving architectures.

\begin{figure*}[t!]
    \centering
    \includegraphics[width=0.8\textwidth]{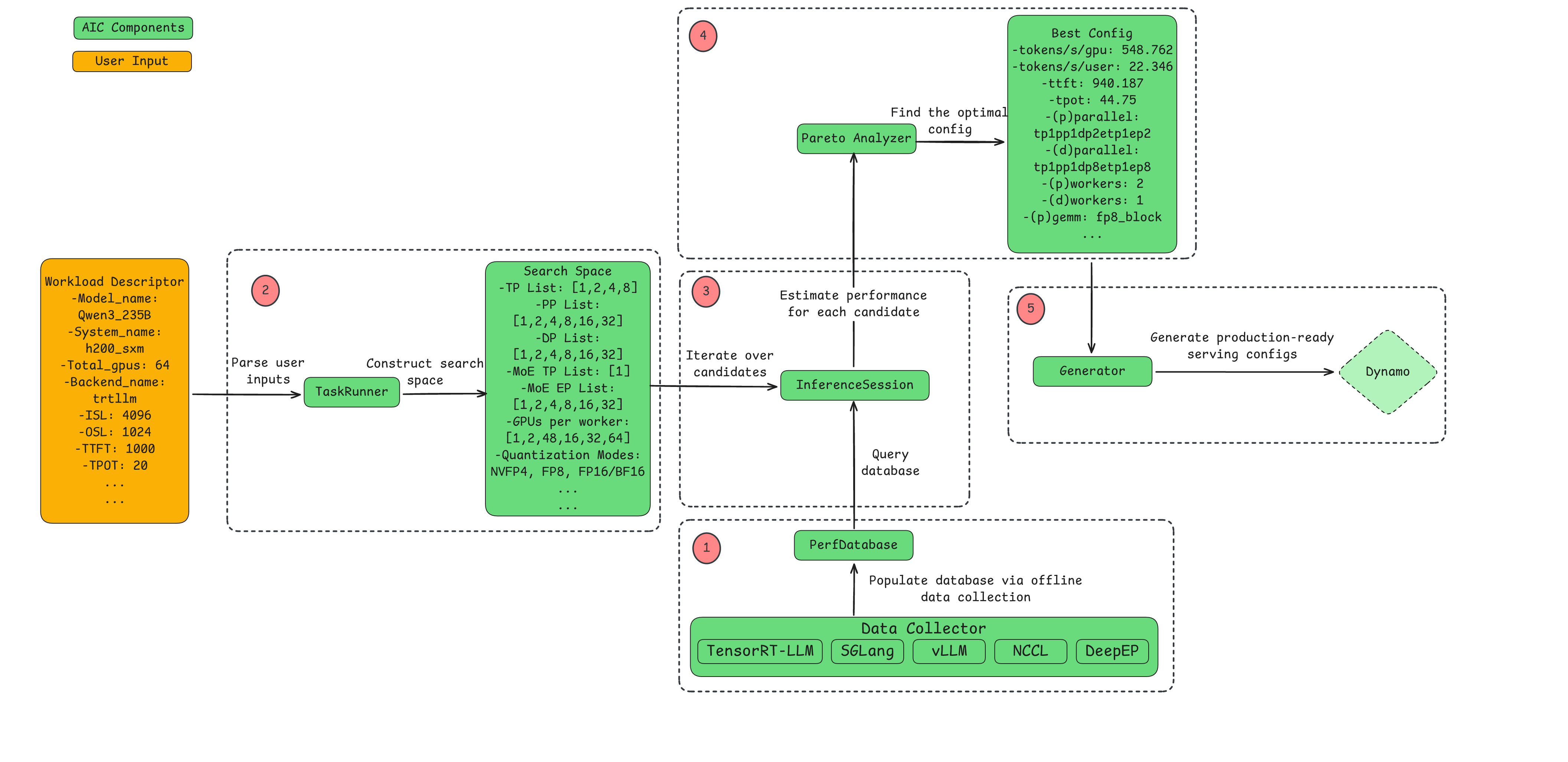}
    \vspace{-2mm}
    \caption{Key components of AIConfigurator and the general workflow of finding the optimal configuration.}
    \label{fig:workflow}
\end{figure*}

\section{Design and Implementation}
AIConfigurator employs a principled, data-driven approach to navigate the vast LLM inference configuration space. Rather than relying on theoretical roofline models or exhaustive benchmarking, the system decomposes inference into fundamental operations, collects real hardware measurements for these primitives, and composes end-to-end performance estimates through a well-defined performance model. 

The toolkit supports multiple inference frameworks through a unified backend abstraction. Each backend (TensorRT-LLM, SGLang, vLLM) implements framework-specific logic for memory estimation, aggregated serving simulation, and constraint-based optimization, while sharing the common operation modeling infrastructure.

This section illustrates the architecture, core mechanisms, and implementation details of AIConfigurator.



\begin{figure*}[t!]
    \centering
    \includegraphics[width=0.75\textwidth]{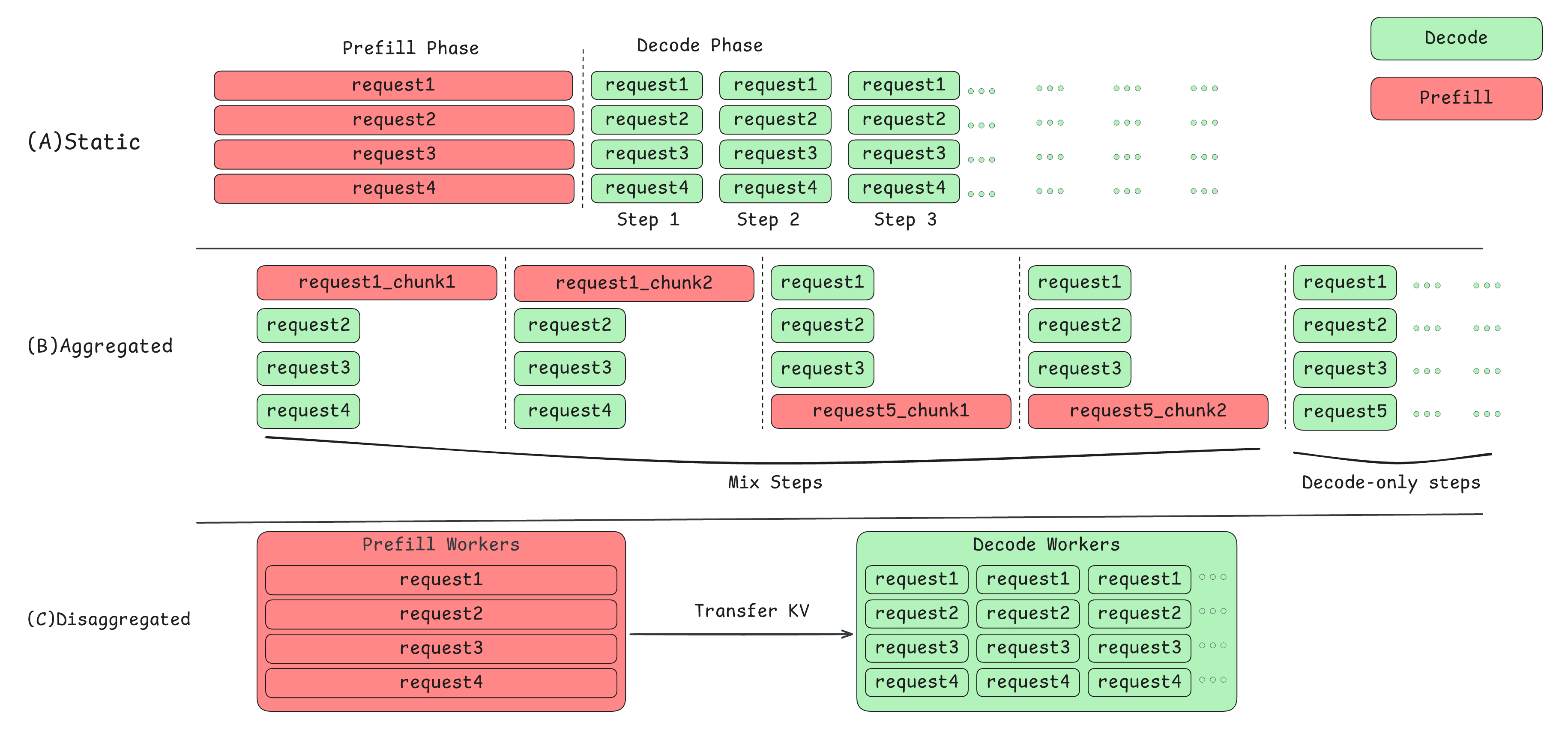}
    \vspace{-2mm}
    \caption{Three serving modes modeled by AIConfigurator. (A) \textbf{Static}: GPU workers process fixed inference requests end-to-end. (B) \textbf{Aggregated}: prefill and decode of different requests are mixed (continuous batching). (C) \textbf{Disaggregated}: separate GPU pools for prefill and decode phases.}
    \label{fig:serving_mode}
\end{figure*}

\subsection{Workflow of AIConfigurator}
Figure~\ref{fig:pareto_chart} shows Pareto curves comparing two serving modes---\textbf{aggregated} and \textbf{disaggregated}---for a \textbf{Qwen-235B} model on 64 \textbf{H200} GPUs. This represents one of the key insights from AIConfigurator's simulation results. The horizontal axis shows \textbf{generation speed} (tokens generated per second per user request), while the vertical axis shows \textbf{system throughput} (tokens generated per second per GPU). Each point represents a serving configuration that satisfies the TTFT constraint.

An optimal configuration is one that achieves the highest system throughput while meeting a target generation speed. For example, with an input sequence length of 4096 and output length of 1024, if we require at least 20 tokens/s per user, the starred configurations are optimal as they maximize per-GPU throughput while exceeding this speed threshold. Notably, disaggregated serving is preferable here: its best configuration achieves 823 tokens/s/GPU, approximately \textbf{53\%} higher than the best aggregated configuration (564 tokens/s/GPU) under the same speed constraint.

In summary, AIConfigurator identifies the optimal serving configuration that maximizes system throughput while meeting specific SLA targets (e.g., TTFT < 1s, generation speed > 20 tokens/s/user).

Figure~\ref{fig:workflow} describes the general workflow of utilizing AIConfigurator, which typically involves five steps, each revolving around one of the key components of AIConfigurator:
\begin{itemize}
    \item \textbf{PerfDatabase}: First, an offline data collection process is performed to construct a database with comprehensive performance data for a wide range of commonly used LLM operators across different designated hardware platforms. Each of the supported inference frameworks is handled independently in this process.
    \item \textbf{TaskRunner}: At the second step, the TaskRunner will construct a search space comprised of all the valid candidate serving configurations based on the user provided workload descriptor, which will include information like user desired environment setup, SLAs and request specific characteristics.
    \item \textbf{InferenceSession}: InferenceSession will then iterate over all the candidate serving configurations, and for each candidate, it will estimate key performance metrics by combining iteration-level modeling with operator-level performance data queried directly from the PerfDatabase.
    \item \textbf{Pareto Analyzer}: Afterwards, the Pareto analyzer will filter and rank all the valid serving configurations based on the performance projections generated in the previous stage, and output the top-ranked configurations with comprehensive performance projections.
    \item \textbf{Generator}: Finally, AIConfigurator's generator module can directly convert the serving recommendations identified by the Pareto analyzer into version compatible launch file for any one of the inference engines among TensorRT-LLM, vLLM and SGLang, automatically setting the optimal serving flags such as \texttt{-{}-enable\_cuda\_graph}, \texttt{-{}-kv\_cache\_free\_gpu\_mem\_fraction} and \texttt{-{}-enable\_chunked\_context}. Serving frameworks like NVIDIA Dynamo can also directly leverage the launch file to set up an optimally configured inference server.
\end{itemize}

\subsection{Performance Modeling}
Instead of directly estimating the system throughput and generation speed, AIConfigurator derives these two metrics from the perspective of an average request arriving at the server operating at a steady state of concurrency.

\begin{equation}
    \text{Generation Speed} = 1000 / TPOT
    \label{eq:loss}
\end{equation}
    \vspace{-2mm}
\begin{equation}
    \begin{split}
    \text{System Throughput} &= \frac{1000}{TTFT + (OSL - 1) \times TPOT} \\
    &* \text{Batch Size} \\
    &* OSL \\ 
    &/ \text{Total Number of GPU}
    \end{split}
    \label{eq:throughput}
\end{equation}

Both \textbf{TTFT}(Time to First Token) and \textbf{TPOT}(Time Per Output Token) are measured in milliseconds. \textbf{OSL}(output sequence length) is treated as a fixed value provided by the user-supplied workload descriptor. As for \textbf{Batch size}, AIConfigurator will sweep over a range of pre-defined values when estimating the performance for a given serving configuration.

In order to estimate TPOT and TTFT efficiently, we first consider the full life cycle of a typical request being processed by an inference server. In the context of LLM serving, any given request will go through two distinct processing phases:

\textbf{The prefill/context phase} processes the entire user input prompt, computing and storing the associated KV cache before producing the first output token. This phase is normally compute intensive, and fused multi-head attention style kernels like FlashAttention~\cite{dao2022flashattention} are usually utilized in this phase to accelerate the attention computation. Additionally, context chunking~\cite{agrawal2024taming} can also be employed to split prefill of long input prompts into multiple iteration steps.

\textbf{The decode/generation phase} generates the remaining output tokens one at a time in a step-by-step auto-regressive fashion. Newly generated tokens are utilized as the input for the subsequent step, and cached key-value pairs are used to avoid recomputing key-value pairs for the past tokens. This phase is normally memory intensive and specialized kernels like XQA~\cite{githubGitHubNVIDIATensorRTLLM} are utilized in this phase.

While prefill and decode constitute the fundamental computational stages of LLM inference, real-world performance depends heavily on how the serving engine orchestrates them. A naive abstraction is insufficient to capture the nuances of different scheduling strategies. Therefore, AIConfigurator explicitly models three distinct serving modes—Static, Aggregated, and Disaggregated—each of which handles resource contention and phase interleaving differently. We first define the baseline behavior under the static mode.
\begin{algorithm}[H]
\caption{Static Mode Inference Performance Estimation}
\label{alg:static_mode}
\begin{algorithmic}[1]

\Require $ISL$ (Input Length), $OSL$ (Output Length)
\Require $B$ (Batch Size), $P$ (Prefix Length)
\Require Stride $S_{stride}$ (Default: 32)
\Require Function $\textsc{GetStepLatency}(batch\_size,$
\Statex \hspace{4.5cm} $seq\_len, phase)$

\State \textbf{Phase 1: Context Latency (TTFT)}
\State $ISL_{eff} \gets ISL - P$
\State \Comment{Calculate latency for processing the input prompt}
\State $\text{TTFT} \gets \textsc{GetStepLatency}(B, ISL_{eff}, \text{'prefill'})$

\State \textbf{Phase 2: Generation Latency (Total Generation Time)}
\State $T_{gen} \gets 0$
\If{$OSL > 1$}
    \State $k \gets 0$
    \While{$k < OSL - 1$}
        \State $S_{seq} \gets ISL + k + 1$ \Comment{Current total sequence length}
        
        \State \Comment{Query latency for a single decode step at current length}
        \State $T_{step} \gets \textsc{GetStepLatency}(B, S_{seq}, \text{'decode'})$
        
        \State $R \gets \min(S_{stride}, OSL - 1 - k)$ \Comment{Interpolate for next R tokens} \label{step:stride}
        \State $T_{gen} \gets T_{gen} + (T_{step} \times R)$
        \State $k \gets k + S_{stride}$
    \EndWhile
\EndIf

\State \textbf{Phase 3: TPOT Calculation}
\If{$OSL > 1$}
    \State $\text{TPOT} \gets T_{gen} / (OSL - 1)$
\Else
    \State $\text{TPOT} \gets 0$
\EndIf
\State \Return $\text{TTFT}, \text{TPOT}$
\end{algorithmic}
\end{algorithm}
\subsubsection{Static Mode}
In the static serving mode, as depicted in Figure~\ref{fig:serving_mode}(A), the workload is processed in a strictly sequential manner with a fixed batch size. In this regime, TTFT is equivalent to the latency of the prefill phase. TPOT is approximated by averaging the latency of the subsequent decoding steps required to generate the entirety of the output sequence.

Algorithm \ref{alg:static_mode} outlines the procedure for calculating these metrics. It employs a stride-based optimization (Step \ref{step:stride}) to reduce the computational overhead of estimating generation latency, interpolating costs over intervals rather than querying the database for every token step. 

\subsubsection{Aggregated Mode}
Also known as inflight batching or continuous batching~\cite{yu2022orca}, this serving mode is distinguished by its ability to mix prefill and decode steps from different requests within a single inference iteration, as shown in Figure \ref{fig:serving_mode}(B). This flexibility significantly improves GPU resource utilization compared to static serving, leading to higher system throughput.

Our performance model, detailed in Algorithm \ref{alg:aggregated_mode}, approximates this behavior by dividing execution into two distinct stages:

\textbf{Mixed Phase:} This phase represents the steady-state operation of a continuous batching system where prefill and decode requests run concurrently. The scheduler prioritizes utilizing the available context capacity ($C_{ctx}$) to process prefill requests. By default a prefill request will contain a ISL number of tokens, and context chunking can be optionally enabled to split full ISL number of context tokens into multiple prefill requests. The remaining batch slots are allocated to decoding tasks ($N_{mix}^{gen}$). Crucially, when the prefill workload is heavy (context processing time exceeds generation time), our model (Algorithm \ref{alg:aggregated_mode}, lines 6-10) throttles the number of concurrent decode streams using a rate-matching heuristic. This prevents the "starvation" scenario where decode requests finish faster than new requests can be prefilled. The step latency ($L_{mix}$) in this phase is dominated by the compute-intensive prefill attention.

\textbf{Generation-Only Phase:} This phase models the tail end of a workload or periods of low arrival rate where the prefill queue has been drained. The system transitions to a pure decoding regime, dedicating all batch slots ($B$) to autoregressive generation. The step latency ($L_{gen}$) here is significantly lower, typically bounded by memory bandwidth rather than compute.

We estimate TTFT based on the latency of the Mixed Phase, incorporating an empirical correction factor ($F_{corr}$) modeled as a piecewise linear function. This factor accounts for base scheduling overhead (constant term), proportional queuing delay as the context backlog grows, and a saturation limit to reflect system-level admission controls. For TPOT, we employ a weighted average of latencies from both phases. Notably, we apply a small offset (3 steps) to the mixed phase duration when calculating weights (Algorithm \ref{alg:aggregated_mode}, Step 5). This heuristic filters out scheduling jitter often observed in the initial steps of a batch, providing a more robust estimate of steady-state decoding performance. This approach models the non-linear interference between prefill and decode operations without requiring a full discrete-event simulation.

\vspace{-1.5em}
\begin{algorithm}[H]
\caption{Aggregated Mode (Continuous Batching) Performance Estimation}
\label{alg:aggregated_mode}
\begin{algorithmic}[1]
\Require $ISL, OSL$
\Require $B$ (Batch Size), $C_{ctx}$ (Context Token Capacity)
\Require Helper Functions:
\Statex \quad $\textsc{GetMixLat}(N_{ctx}, N_{gen}, ISL, OSL)$
\Statex \quad $\textsc{GetGenLat}(N_{gen}, ISL, OSL)$

\State \textbf{Step 1: Phase Duration (in Steps)}
\State $T_{total\_ctx} \gets \lceil (ISL \times B) / C_{ctx} \rceil$ \Comment{Total steps to process all context}

\State \textbf{Step 2: Workload Distribution (Steps \& Tokens)}
\If{$B > 1$}
    \If{$T_{total\_ctx} \geq OSL$}
        \State \Comment{Context dominates; generation slots are limited}
        \State $T_{mix} \gets T_{total\_ctx}$
        \State $T_{gen} \gets 0$
        \State $N_{mix}^{ctx} \gets C_{ctx}$ \Comment{Tokens per step}
        \State $N_{mix}^{gen} \gets \max(1, \lfloor B / (T_{total\_ctx} / OSL) \rfloor)$ \Comment{Tokens per step}
    \Else
        \State \Comment{Standard continuous batching}
        \State $T_{mix} \gets T_{total\_ctx}$
        \State $T_{gen} \gets OSL - T_{mix}$
        \State $N_{mix}^{ctx} \gets C_{ctx}$
        \State $N_{mix}^{gen} \gets B - \lceil C_{ctx} / ISL \rceil$ \Comment{Fill remaining slots}
        \State \textbf{assert} $N_{mix}^{gen} \ge 1$
    \EndIf
\Else
    \State \Comment{Single Batch ($B=1$)}
    \State $T_{mix} \gets 1, \quad T_{gen} \gets OSL - 1$
    \State $N_{mix}^{ctx} \gets C_{ctx}, \quad N_{mix}^{gen} \gets 0$
\EndIf

\State \textbf{Step 3: Latency Calculation}
\State $L_{mix} \gets \textsc{GetMixLat}(N_{mix}^{ctx}, N_{mix}^{gen}, ISL, OSL)$
\State $L_{gen} \gets \textsc{GetGenLat}(B, ISL, OSL)$

\State \textbf{Step 4: TTFT Estimation}
\State $F_{corr} \gets \min\left(2 + \frac{T_{total\_ctx} - 3}{20}, 4\right)$ 
\State $\text{TTFT} \gets L_{mix} \times \lceil ISL / C_{ctx} \rceil \times F_{corr}$

\State \textbf{Step 5: TPOT Estimation}
\State $T'_{mix} \gets \max(1, T_{mix} - 3)$ 
\If{$B > 1$}
    \State $\text{TPOT} \gets \frac{L_{mix} \times T'_{mix} + L_{gen} \times T_{gen}}{T'_{mix} + T_{gen}}$
\Else
    \State $\text{TPOT} \gets L_{gen}$
\EndIf

\State \Return $\text{TTFT}, \text{TPOT}$
\end{algorithmic}
\end{algorithm}

\subsubsection{Disaggregated Mode}
\begin{algorithm}[H]
\caption{Disaggregated Mode Performance Estimation}
\label{alg:disagg_mode}
\begin{algorithmic}[1]
\Require Candidate Configs: $C_{pre}$ (Prefill), $C_{dec}$ (Decode)
\Require Constraints: Max TTFT ($L_{limit}^{TTFT}$), Max TPOT ($L_{limit}^{TPOT}$)
\Require Valid Total GPU Counts: $G_{valid}$
\Require Degradation Factors: $\alpha_{pre} = 0.9, \alpha_{dec} = 0.92$
\Require TTFT Correction Factor: $\beta_{TTFT} = 1.8$

\State \textbf{Step 1: Filter Candidates by Latency}
\State \Comment{Filter prefill configs ($c_{pre}$) and decode configs ($c_{dec}$)}
\State $C'_{pre} \gets \{ c \in C_{pre} \mid (L_{pre}(c) \times \beta_{TTFT}) \le L_{limit}^{TTFT} \}$
\State $C'_{dec} \gets \{ c \in C_{dec} \mid L_{dec}(c) \le L_{limit}^{TPOT} \}$

\State \textbf{Step 2: Rate Matching (Find Optimal $x, y$)}
\State $BestConfig \gets \emptyset, \quad MaxThru_{GPU} \gets 0$

\For{$c_{dec} \in C'_{dec}$}
    \For{$c_{pre} \in C'_{pre}$}
        \State $Thru_{pre} \gets \text{SeqThroughput}(c_{pre})$
        \State $Thru_{dec} \gets \text{SeqThroughput}(c_{dec})$
        \State $G_{pre} \gets \text{GPUs}(c_{pre}), \quad G_{dec} \gets \text{GPUs}(c_{dec})$

        \State \Comment{Sweep worker counts $x$ (prefill) and $y$ (decode)}
        \For{$x \in [1, 32], \ y \in [1, 64]$}
            \State $G_{total} \gets x \cdot G_{pre} + y \cdot G_{dec}$
            \If{$G_{total} \notin G_{valid}$} \textbf{continue} \EndIf

            \State $R_{pre} \gets Thru_{pre} \cdot x \cdot \alpha_{pre}$
            \State $R_{dec} \gets Thru_{dec} \cdot y \cdot \alpha_{dec}$
            
            \State $R_{sys} \gets \min(R_{pre}, R_{dec})$ \Comment{System Rate}
            \State $Thru_{GPU} \gets R_{sys} / G_{total}$

            \If{$Thru_{GPU} > MaxThru_{GPU}$}
                \State $MaxThru_{GPU} \gets Thru_{GPU}$
                \State $BestConfig \gets \{ \text{TTFT}: L_{pre}(c_{pre}), \text{TPOT}: L_{dec}(c_{dec}), \dots \}$
            \EndIf
        \EndFor
    \EndFor
\EndFor

\State \Return $BestConfig$
\end{algorithmic}
\end{algorithm}
In contrast to the static and aggregated modes, the Disaggregated Mode employs two distinct worker pools, each dedicated to a specific phase of LLM inference. As illustrated in Figure~\ref{fig:serving_mode}(C), incoming requests are first processed by dedicated prefill workers. Upon completion of the prefill phase, the computed key-value (KV) cache and intermediate states are transferred to decode workers, which generate subsequent tokens in an autoregressive manner.

This decoupling offers significant architectural advantages: it eliminates prefill-decode interference and allows each pool to employ distinct model parallelism strategies specialized for its specific workload characteristics—optimizing compute-bound prefill and memory-bound decoding independently. This separation has been shown to significantly improve overall system Goodput in prior studies~\cite{zhong2024distservedisaggregatingprefilldecoding}.

From a modeling perspective, AIConfigurator estimates the performance of a disaggregated system in two stages. Firstly, it independently sweeps the search space for prefill and decode configurations, treating each candidate as an isolated static instance and estimating its base latency using Algorithm \ref{alg:static_mode}. It applies a correction factor ($\beta_{TTFT}$) to the prefill latency to account for the KV-cache transmission overhead inherent to disaggregated architectures.

Secondly, it constructs valid composite servers—denoted as $(x)P(y)D$, where $x$ and $y$ represent the number of prefill and decode instances, respectively—through a \textbf{rate-matching} process. The algorithm identifies the optimal configuration by maximizing the effective \textbf{per-GPU throughput} ($Thru_{GPU}$), derived from the system rate $R_{sys}$:
$$ R_{sys} = \min(R_{pre}, R_{dec}) $$
where $R$ represents the request rate (requests per second) of the respective worker pools, discounted by interference factors $\alpha$. Once the optimal $(x)P(y)D$ configuration is identified, the system's \textbf{TTFT} is derived from the latency of the selected prefill workers pool (including transfer overhead), while \textbf{TPOT} is determined by the latency of the decode workers pool. Algorithm \ref{alg:disagg_mode} details this optimization procedure.

\subsection{Iteration-level Modeling}
The effectiveness of all three of the algorithms introduced thus far depends on how accurately we can predict the latency of a given inference iteration step. For instance, in Algorithm \ref{alg:static_mode}, we rely on the latency data of prefill-only step and decode-step acquired via $\textsc{GetStepLatency}(batch\_size, seq\_len, phase)$ to perform further estimation, while in Algorithm \ref{alg:aggregated_mode}, besides decode-only step ($\textsc{GetGenLat}(N_{gen}, ISL, OSL)$), we must also obtain accurate latency measurement for the mixed step ($\textsc{GetMixLat}(N_{ctx}, N_{gen}, ISL, OSL)$) in order to move forward. Therefore, aiming at establishing a fast and robust way of estimating the latency of a given iteration step, AIConfigurator approaches this issue by fully exploiting the decomposable nature of the modern LLM inference.

\subsubsection{Decompose Iteration Into Operators}
As modern LLMs are composed of repetitive transformer layers, any inference iteration step can then be modeled as running a fixed sequence of operators for a number of times, typically depending on the number of layers that the model possesses. Introducing modern parallel strategies does not alter this fundamental property except for inserting a few well-defined communication operators at fixed positions of the iteration's execution path and scaling down the compute operators by sharding inputs across multiple compute devices.

\begin{figure}[H]
    \centering
    \includegraphics[width=1.0\linewidth]{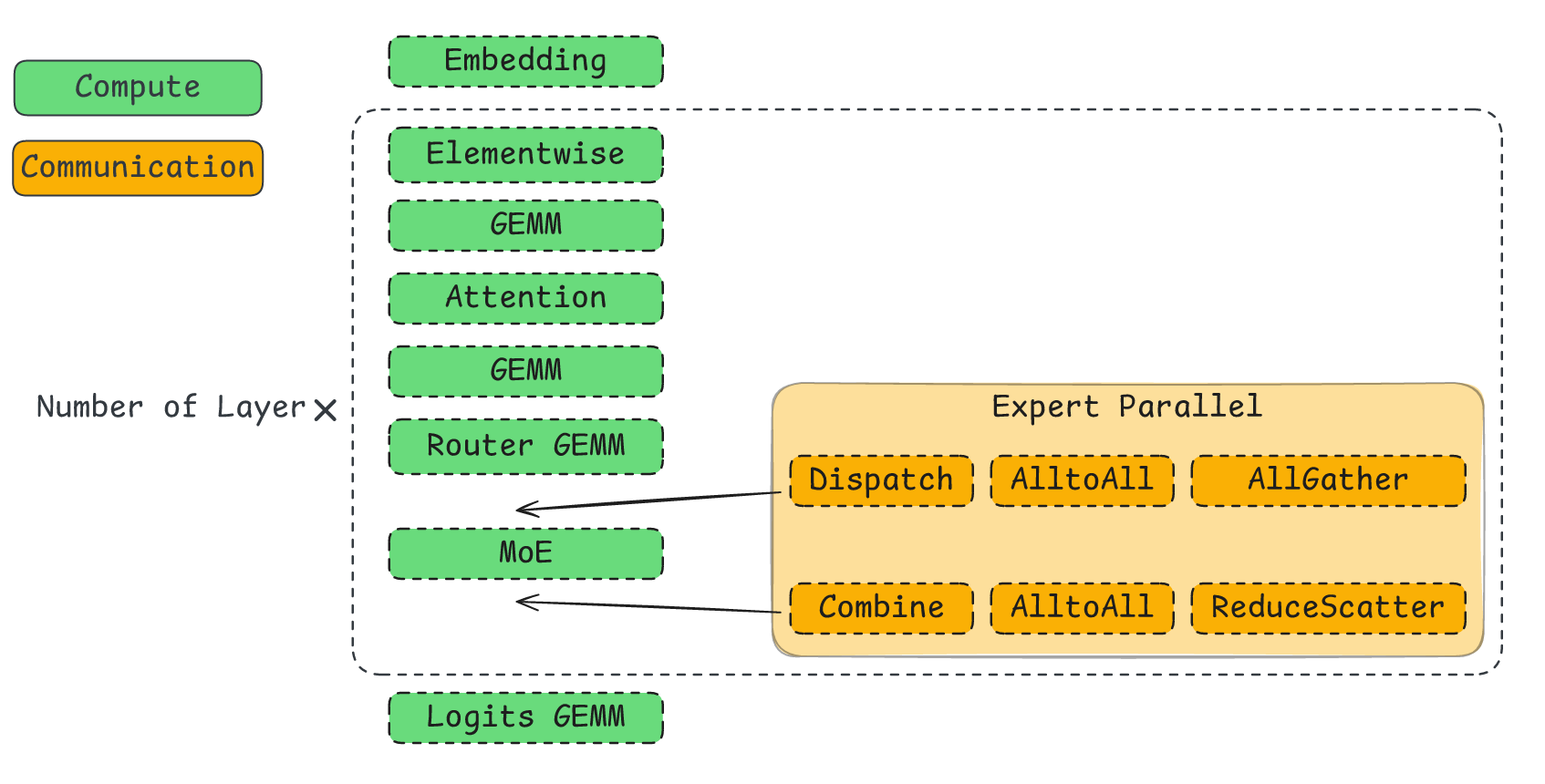}
       \vspace{-5.5mm}
    \caption{\small A step of LLM inference can be decomposed into repeated execution of a few key operators. For instance, the inference step of a typical MoE model normally involves the above depicted operators, and how expert parallelism is implemented depends on the specific backend used.}
    \label{fig:moe_model}
\end{figure}

For instance, Figure~\ref{fig:moe_model} depicts a typical composition of operators while performing inference with an MoE model. The entire inference step essentially amounts to running 4 types of operators, namely embedding, GEMM, Attention(prefill or decode, depending on the phase) and MoE for a number of times. If expert parallelism is added to the mix, it will scale down the size of MoE operator while adding a pair of communication operators, and the exact pair shall depend on the inference engine backend used in production. 

Given the above observations, in AIConfigurator, we model the latency of an inference iteration step by aggregating the performance profiles of its constituent operators.
\subsection{Operator Database}
AIConfigurator constructs a performance database through offline profiling on actual GPU hardware, currently supporting TensorRT-LLM, vLLM, and SGLang.

\textbf{Database Coverage.} The database includes: (1) \emph{GEMM operations} parameterized by dimensions $(M, N, K)$ and quantization (FP16, FP8, INT8, INT4); (2) \emph{Attention operations} for both compute-bound context attention and memory-bound generation attention, supporting MHA~\cite{vaswani2023attentionneed}, GQA~\cite{ainslie2023gqatraininggeneralizedmultiquery}, and MLA~\cite{deepseekai2024deepseekv2strongeconomicalefficient}; (3) \emph{Communication primitives} including AllReduce, AllGather, AllToAll, and point-to-point transfers across message sizes and GPU counts; (4) MoE operations with \emph{dispatch/combine}~\cite{deepep2025} patterns; and (5) \emph{Hardware specifications} (memory bandwidth, compute throughput, interconnect bandwidth).

\textbf{Data Collection.} We combine three strategies: \emph{exhaustive profiling} sweeps parameters (batch size, sequence length, hidden dimension) with framework-native tools ($\sim$30 GPU-hours per platform-framework pair); \emph{interpolation} estimates latencies for intermediate configurations using profiled data points; and \emph{Speed-of-Light estimation} provides analytical bounds via roofline models for unprofiled operators.

\subsubsection{Power Law Correction for MoE}
\begin{figure}[ht]
    \centering
    \begin{tikzpicture}
        \begin{axis}[
            width=\linewidth,
            height=6cm,
            ybar,
            bar width=7pt,
            ylabel={Normalized Load},
            xlabel={Experts (Ranked by Usage)},
            symbolic x coords={E1, E2, E3, E4, E5, E6, E7, E8},
            xtick=data,
            ymin=0,
            legend style={at={(0.95,0.95)}, anchor=north east},
            grid=major,
            title={Effect of ($\alpha$) on Expert Load Balancing}
        ]
            \addplot coordinates {
                (E1,0.125) (E2,0.125) (E3,0.125) (E4,0.125) 
                (E5,0.125) (E6,0.125) (E7,0.125) (E8,0.125)
            };

            \addplot coordinates {
                (E1,0.34) (E2,0.17) (E3,0.11) (E4,0.08) 
                (E5,0.07) (E6,0.06) (E7,0.05) (E8,0.04)
            };

            \addplot coordinates {
                (E1,0.45) (E2,0.19) (E3,0.12) (E4,0.08) 
                (E5,0.06) (E6,0.04) (E7,0.03) (E8,0.02)
            };

            \legend{$\alpha=0.0$ (Balanced), $\alpha=1.01$ (Zipfian), $\alpha=1.2$ (High Skew)}
        \end{axis}
    \end{tikzpicture}
    \caption{\small Visualizing the effect of $\alpha$. As $\alpha$ increases, the routing distribution shifts from perfectly balanced (uniform) to highly skewed, where the top-ranked experts (E1, E2) handle the majority of tokens.}
    \label{fig:expert_load_distribution}
\end{figure}
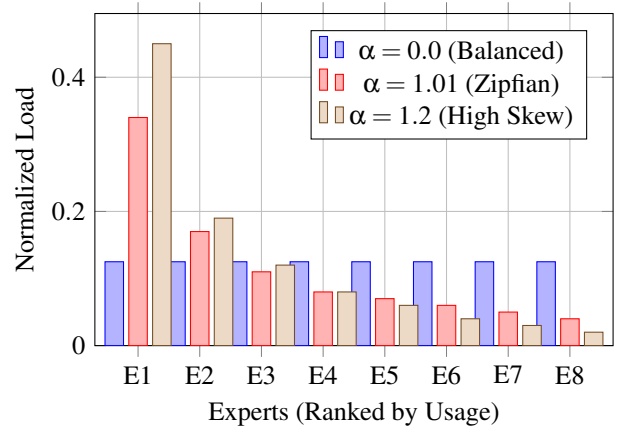

MoE operator performance depends heavily on token distribution. Prior works~\cite{fedus2022switch, lepikhin2021gshard} show that during inference, certain experts receive disproportionately more tokens---observations from Qwen3-235B indicate $\sim$70\% of compute is handled by only 20\% of active experts. To account for this imbalance, AIConfigurator implements a controlled token assignment procedure that emulates power-law distributions observed in production (Figure~\ref{fig:expert_load_distribution}).

\textbf{Step 1: Sample Expert Load Weights.}
We generate a load profile by sampling $E$ weights (one per expert) from a power-law distribution. Using inverse transform sampling with $U \sim \text{Uniform}(0, 1)$, each raw weight is computed as:
\begin{equation}
x_i = \left[ (x_{\max}^{1-\alpha} - x_{\min}^{1-\alpha}) \cdot U + x_{\min}^{1-\alpha} \right]^{\frac{1}{1-\alpha}}
\end{equation}
where $x_{\min}$ and $x_{\max}$ define the distribution bounds, and $\alpha$ controls the degree of imbalance. These weights are then normalized to obtain the token count for each expert:
\begin{equation}
N_i = \text{round}\left( \frac{x_i}{\sum_{j=0}^{E-1} x_j} \times T_{\text{total}} \times K \right)
\end{equation}
where $T_{\text{total}}$ is the batch size, $K$ is the top-k routing factor (each token routes to $K$ experts), and $N_i$ is the number of tokens assigned to expert $i$. Residual tokens from rounding are distributed to balance the total. The parameter $\alpha$ controls the skew: $\alpha \approx 0$ yields nearly uniform load (theoretical ideal), while $\alpha \approx 1.2$ produces heavy-tailed distributions where a few ``hot'' experts receive most tokens---matching observations from models like Qwen3-235B.

\textbf{Step 2: Construct Synthetic Router Assignments.}
During normal inference, a learned gating network routes each token to its assigned expert(s). For controlled benchmarking, we bypass this router and directly inject a synthetic assignment matrix $\mathbf{L} \in \mathbb{R}^{T_{\text{total}} \times E}$, where exactly $N_i$ tokens are deterministically routed to expert $i$. This eliminates stochastic variance from the router and ensures the hardware executes the precise workload shape from Step~1---allowing us to capture the "tail latency" caused by the most heavily loaded expert, which determines overall throughput in practice.

\section{Evaluation}
We evaluate AIConfigurator along three dimensions: (1) prediction fidelity against ground-truth hardware measurements, (2) search efficiency compared to exhaustive benchmarking, and (3) practical performance gains in production scenarios. Each subsection describes its own experimental setup.

\subsection{Aggregated Serving Fidelity}

We first evaluate AIConfigurator's prediction accuracy for aggregated (continuous batching) serving, the most common deployment mode in production LLM inference.

\textbf{Setup.}
Experiments were conducted on a single NVIDIA H100 SXM node with 8 GPUs (80GB HBM3 each) connected via NVSwitch. We evaluated two models spanning different sizes and architectures: \textbf{Qwen3-32B} (32B parameters, dense, FP8) and \textbf{Qwen3-235B} (235B parameters, Mixture-of-Experts with 128 experts, FP8). All models were served using TensorRT-LLM v1.0.0. To assess cross-framework generalization, we also evaluated \textbf{Qwen3-32B} using \textbf{vLLM v0.11.0}.

We swept a comprehensive configuration space to stress-test prediction accuracy across diverse operating points:
\begin{itemize}[noitemsep,topsep=0pt]
    \item \textbf{Input Sequence Length (ISL):} 128--4096 tokens
    \item \textbf{Output Sequence Length (OSL):} 128--512 tokens
    \item \textbf{Concurrency:} 4--128 concurrent requests
    \item \textbf{Tensor Parallelism (TP):} 1, 2, 4, 8 GPUs
    \item \textbf{Expert Parallelism (EP):} 1, 2, 4, 8 GPUs (for Qwen3-235B)
\end{itemize}
This yields \textbf{960 unique configurations} for TensorRT-LLM (360 for Qwen3-32B and 600 for Qwen3-235B) plus 128 configurations for vLLM, covering workloads from latency-sensitive chat to throughput-oriented batch processing.

\textbf{Metrics.}
We measure prediction fidelity using Mean Absolute Percentage Error (MAPE) for two key latency metrics: Time-Per-Output-Token (TPOT), which determines generation throughput, and Time-To-First-Token (TTFT), which governs user-perceived responsiveness.

\begin{figure}[t]
    \centering
    \includegraphics[width=\linewidth]{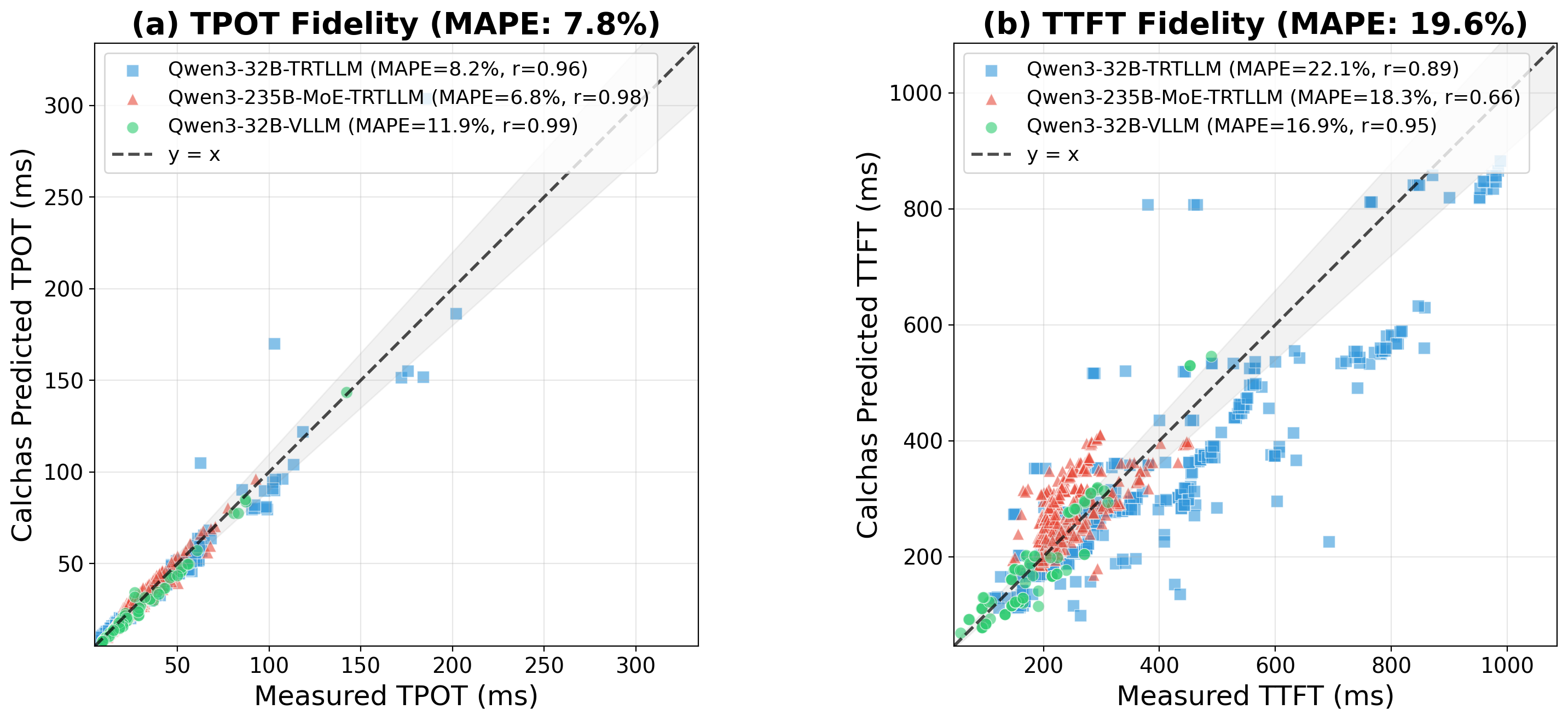}
    \caption{Prediction fidelity for aggregated serving across TensorRT-LLM and vLLM on H100 SXM. Each point represents one configuration; the diagonal indicates perfect prediction. TTFT values $>$ 1000ms are filtered as outliers.}
    \label{fig:agg_fidelity}
\end{figure}

\textbf{Results.}
Figure~\ref{fig:agg_fidelity} presents the fidelity analysis across both frameworks. For TPOT prediction, AIConfigurator achieves an overall MAPE of \textbf{7.8\%} with strong correlation across all models: Qwen3-32B-TRTLLM (8.2\%, $r$=0.96), Qwen3-235B-MoE-TRTLLM (6.8\%, $r$=0.98), and Qwen3-32B-VLLM (11.9\%, $r$=0.99). The MoE model achieves the lowest error, validating our power-law workload modeling for expert load imbalance. The consistent accuracy across TensorRT-LLM and vLLM demonstrates that our operator-level modeling generalizes well---the core computational patterns (GEMM, attention, communication) are predictable regardless of framework.

TTFT prediction also demonstrates strong fidelity: Qwen3-32B-TRTLLM achieves 22.1\% MAPE ($r$=0.89), Qwen3-235B-MoE-TRTLLM achieves 18.3\% MAPE ($r$=0.66), and Qwen3-32B-VLLM achieves \textbf{16.9\%} MAPE ($r$=0.95). Notably, vLLM shows the best TTFT prediction accuracy, likely due to its more predictable prefill scheduling compared to TensorRT-LLM's batching heuristics. Extreme outliers (TTFT $>$ 1000ms) are excluded as they represent pathological queuing delays rather than steady-state operation. The strong correlation across both metrics confirms AIConfigurator's utility for configuration selection across heterogeneous framework deployments.

\begin{figure*}[t]
    \centering
    \includegraphics[width=1.0\textwidth]{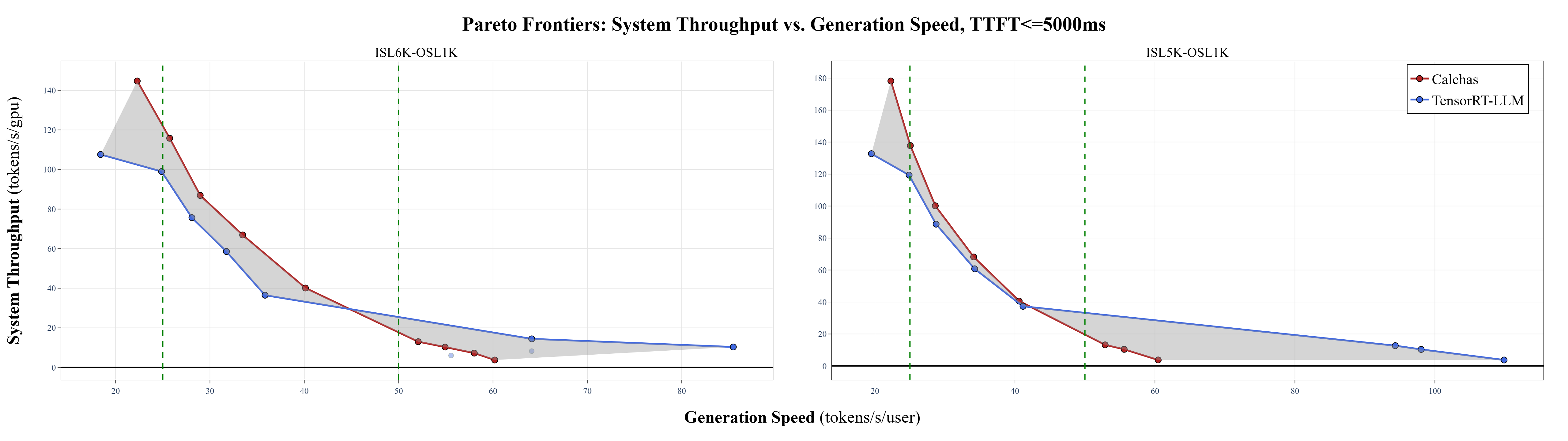}
    \caption{\small AIConfigurator projected Pareto frontier (red) vs.\ TensorRT-LLM ground truth (blue) for DeepSeek V3 deployed across two nodes using prefill/decode disaggregation. Shaded region indicates the discrepancy between AIConfigurator's projections and framework reality.}
    \vspace{-2.5mm}
    \label{fig:aiconfigurator_vs_trtllm}
\end{figure*}

\subsection{Disaggregated Serving Fidelity}

Beyond aggregated mode, we evaluated AIConfigurator's prediction fidelity for multi-node disaggregated serving, where prefill and decode phases run on separate GPU pools.

\textbf{Setup.}
Experiments were conducted on two compute nodes, each equipped with 8 NVIDIA Hopper GPUs interconnected via NVLink. One node was dedicated to prefill while the other handled decode, with parallelism settings configured independently per node. We evaluated the full-size \textbf{DeepSeek V3}~\cite{liu2024deepseek} model (671B parameters, MoE) using \textbf{TensorRT-LLM} as the serving backend.

Unlike aggregated mode where we measure per-request TTFT and TPOT, disaggregated serving requires system-level metrics: \emph{generation speed} (tokens/s/user) and \emph{system throughput} (total tokens/s). We evaluated AIConfigurator in two steps:
\begin{enumerate}[noitemsep,topsep=0pt]
    \item \textbf{Configuration search:} AIConfigurator explored the configuration space for two input profiles (ISL 5k and 6k tokens, OSL 1k tokens) under a 5-second TTFT constraint. The search space included:
    \begin{itemize}[noitemsep,topsep=0pt]
        \item \textbf{Tensor Parallelism (TP):} 1, 2, 4, 8
        \item \textbf{Data Parallelism (DP):} 1, 2, 4, 8
        \item \textbf{Expert Parallelism (EP):} 1, 2, 4, 8
    \end{itemize}
    Configurations exceeding memory capacity were automatically pruned. This sweep produced a Pareto frontier of optimal throughput-vs-speed trade-offs.

    \item \textbf{Ground-truth validation:} We benchmarked each Pareto-optimal configuration on actual TensorRT-LLM and measured MAPE between AIConfigurator predictions and ground-truth measurements.
\end{enumerate}

\textbf{Results.}
Figure~\ref{fig:aiconfigurator_vs_trtllm} compares AIConfigurator projections against TensorRT-LLM measurements. Across all configurations, we observed a MAPE of \textbf{25.49\%} for system throughput and \textbf{14.94\%} for generation speed. Notably, within the interactive speed region of 25--50 tokens/s/user (indicated by dashed green lines), which represents comfortable reading speed for most users, fidelity improves significantly: MAPEs reduce to \textbf{13.19\%} for throughput and \textbf{3.35\%} for generation speed. This demonstrates that AIConfigurator predictions are most accurate precisely where they matter most---the operating region where production deployments typically target.

\subsection{Search Efficiency}


\begin{table}[h]
\centering
\caption{Configuration search efficiency: AIConfigurator vs.\ GPU benchmarking on H100 SXM.}
\label{tab:efficiency}
\footnotesize
\resizebox{\columnwidth}{!}{%
\begin{tabular}{lrrr}
\toprule
\textbf{Model} & \textbf{AIConfigurator} & \textbf{GPU Bench} & \textbf{Speedup} \\
\midrule
\multicolumn{4}{l}{\textit{Total time (all configurations)}} \\
Llama3.1-8B (339 configs)    & 0.52s   & 24.4 hr  & 171,000$\times$ \\
Qwen3-32B FP8 (358 configs)  & 0.72s   & 35.4 hr  & 177,000$\times$ \\
Qwen3-235B FP8 (506 configs) & 0.84s   & 99.5 hr  & 427,000$\times$ \\
\midrule
\multicolumn{4}{l}{\textit{Median time per configuration}} \\
Llama3.1-8B    & 1.5ms  & 4.0 min  & 162,000$\times$ \\
Qwen3-32B FP8  & 1.5ms  & 5.4 min  & 214,000$\times$ \\
Qwen3-235B FP8 & 1.5ms  & 11.5 min & 459,000$\times$ \\
\bottomrule
\end{tabular}%
}
\end{table}

A critical advantage of AIConfigurator is its ability to explore the configuration space on CPU, eliminating the need for exhaustive GPU benchmarking. To quantify this efficiency gain, we compare the time required for AIConfigurator to evaluate the search space of configurations against the wall-clock time needed to benchmark the same configurations.

Table~\ref{tab:efficiency} presents the efficiency comparison across three models of varying complexity. Note that the "GPU time" represents the end-to-end wall-clock time including weight loading, server startup, and benchmark execution. AIConfigurator completes configuration search in sub-second time on CPU, while equivalent GPU benchmarking requires days of GPU time. For example, for Qwen3-235B (506 configurations), AIConfigurator achieves a 427,000$\times$ speedup (0.8s vs.\ 99.5 GPU-hours).

Notably, median per-configuration simulation time remains constant at $\sim$1.5ms regardless of model size, as AIConfigurator's operator-level database queries scale with model architecture rather than parameter count. In contrast, per-configuration benchmarking time grows with model complexity due to longer weight loading and inference times, ranging from 4 to 11.5 minutes.

This efficiency enables practitioners to rapidly iterate on deployment scenarios---adjusting SLA targets, exploring different hardware allocations, or evaluating new model variants---without incurring prohibitive GPU costs.

\subsection{Case Study: Finding the Optimal}

We demonstrate AIConfigurator's practical value by finding optimal serving configurations for a production deployment scenario, comparing aggregated vs.\ disaggregated serving under realistic SLA constraints.

\begin{figure}[H]
    \centering
    \includegraphics[width=0.5\textwidth]{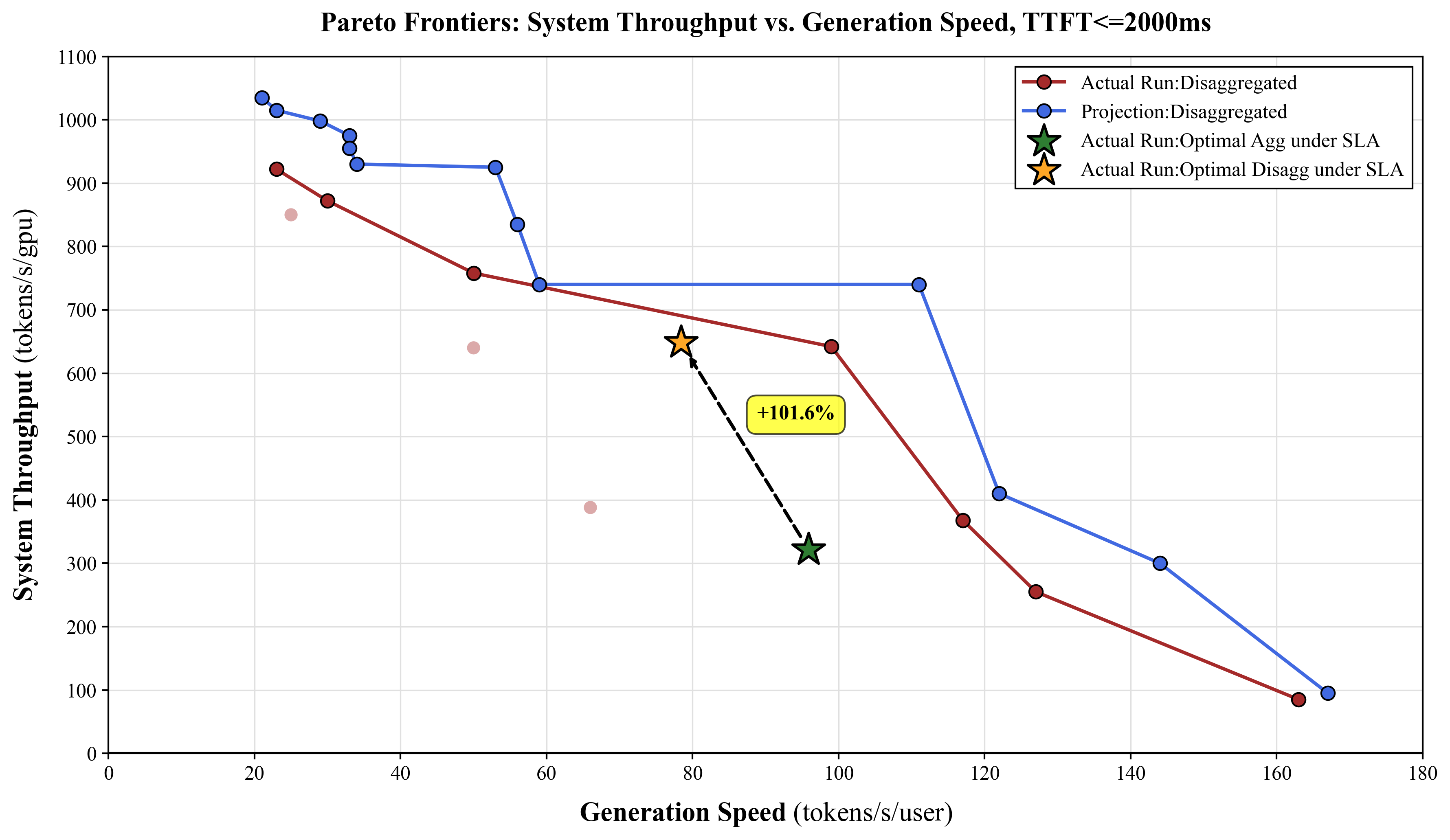}
    \caption{AIConfigurator projections vs.\ ground-truth measurements for Qwen3-32B-FP8 on 8 H200 GPUs. The Pareto frontiers show throughput-vs-speed trade-offs; disaggregated serving achieves 2$\times$ higher throughput than the optimal aggregated configuration while meeting SLA constraints.}
    \label{fig:case_study}
\end{figure}

\textbf{Setup.}
We target the following SLA requirements: TTFT $\leq$ 1200ms, generation speed $\geq$ 60 tokens/s/user, on 8 NVIDIA H200 SXM GPUs. The workload consists of ISL = 4000 and OSL = 500 tokens. All experiments use NVIDIA Dynamo with TensorRT-LLM backend (v0.5.0) and AI-Perf~\cite{aiperf2025} for load testing. Request concurrency matches the maximum batch size to maximize throughput, with 20$\times$ oversampling to mitigate warmup effects on TTFT measurements.

\textbf{Aggregated Baseline.}
Using AIConfigurator, we identified the optimal aggregated configuration: a single TP2 instance with batch size 8, achieving 321.5 tokens/s/GPU and 95.9 tokens/s/user (Table~\ref{tab:agg_disagg_performance_comparison}). We validated this as the global optimum by exhaustively benchmarking all valid TP and batch size combinations.

\textbf{Disaggregated Optimization.}
AIConfigurator explored the disaggregated configuration space in tens of seconds and identified a prefill/decode split: 4 prefill replicas (TP1) and 2 decode replicas (TP2), with batch sizes of 1 and 80 respectively. This configuration achieves \textbf{648.3 tokens/s/GPU}---a \textbf{101.6\% throughput improvement} over the aggregated baseline---while satisfying all SLA constraints (Table~\ref{tab:agg_disagg_performance_comparison}).

\begin{table*}[t]
\centering
\caption{Optimal aggregated vs.\ disaggregated configurations for Qwen3-32B-FP8 on 8 H200 GPUs under production SLA (TTFT $\leq$ 1200ms, speed $\geq$ 60 tokens/s/user). P = prefill, D = decode.}
\label{tab:agg_disagg_performance_comparison}
\resizebox{\textwidth}{!}{
\begin{tabular}{lcccccc}
\toprule
\textbf{Mode} & \textbf{Throughput (tokens/s/GPU)} & \textbf{Speed (tokens/s/user)} & \textbf{TTFT (ms)} & \textbf{Batch Size} & \textbf{Configuration} \\
\midrule
Aggregated & 321.5 & 95.9 & 1017.5 & 8 & 1 $\times$ TP2 \\
Disaggregated & 648.3 & 78.4 & 1068.9 & P:1, D:80 & P: 4 $\times$ TP1, D: 2 $\times$ TP2 \\
\bottomrule
\end{tabular}
}
\end{table*}

\textbf{Projection Accuracy.}
To validate AIConfigurator's predictions across the Pareto frontier, we benchmarked all recommended configurations under a relaxed TTFT constraint of 2000ms (Figure~\ref{fig:case_study}). The AIConfigurator-projected frontier closely tracks ground-truth measurements, with maximum deviations of 11.2\% for generation speed and 17.4\% for system throughput at identical concurrency levels.

\section{Related Work}

\textbf{LLM Inference Systems.} Modern serving has evolved from static batching to dynamic scheduling via Orca~\cite{yu2022orca} (continuous batching), vLLM~\cite{kwon2023efficient} (PagedAttention), SGLang~\cite{zheng2023efficiently} (RadixAttention), and TensorRT-LLM~\cite{githubGitHubNVIDIATensorRTLLM} (optimized kernels). Model parallelism techniques---TP~\cite{shoeybi2019megatron}, PP~\cite{huang2019gpipe}, EP~\cite{liu2024deepseek}---distribute large models across GPUs with complex performance trade-offs. Disaggregated architectures~\cite{zhong2024distservedisaggregatingprefilldecoding, qin2024mooncake, patel2024splitwise} separate prefill/decode phases for independent scaling. AIConfigurator targets these systems, automating configuration selection rather than proposing new scheduling algorithms.

\textbf{Performance Simulation.} Vidur~\cite{agrawal2024vidur} and APEX~\cite{lin2024apex} enable rapid configuration exploration via discrete-event simulation and cost optimization. However, these rely on analytical roofline models that abstract framework-specific behavior. AIConfigurator differs through its data-driven foundation: measuring actual execution times on target hardware and composing these measurements to predict end-to-end performance, capturing implementation-specific overheads that analytical models miss. Unlike static resources like InferenceMax~\cite{semianalysisInferenceMAXSemiAnalysis} benchmarks and deployment recipes~\cite{githubGitHubVllmprojectrecipes}, AIConfigurator provides algorithmic search that generalizes across novel workload combinations.

\section{Conclusion}

We presented AIConfigurator, a data-driven toolkit for optimizing LLM inference configurations across TensorRT-LLM, vLLM, and SGLang. By decomposing inference into fundamental operations and measuring their latencies on target hardware, AIConfigurator achieves high-fidelity performance predictions that capture framework-specific overheads missed by analytical simulators. The toolkit evaluates thousands of configurations in seconds on CPU, eliminating expensive GPU benchmarking campaigns.

Our evaluation demonstrates strong prediction accuracy (6--12\% MAPE for TPOT) across dense and MoE architectures, with production case studies showing 2$\times$ throughput improvements via automated disaggregated configuration discovery. Integration with NVIDIA Dynamo also enables seamless deployment of optimized configurations. Future work includes extending to additional hardware platforms, incorporating cost models, and supporting emerging techniques like speculative decoding and sparse attention.
\clearpage
{\footnotesize \bibliographystyle{acm}
\bibliography{sample}}

\end{document}